\documentclass[11pt,a4paper]{article}

\usepackage[
    a4paper,
    top=0.85in,
    bottom=0.90in,
    left=0.85in,
    right=0.85in
]{geometry}

\usepackage[T1]{fontenc}
\usepackage[utf8]{inputenc}
\usepackage{times}
\usepackage{microtype}
\usepackage{url}
\usepackage{xcolor}

\usepackage{amsmath}
\usepackage{amssymb}
\usepackage{latexsym}
\allowdisplaybreaks

\usepackage{graphicx}
\usepackage{booktabs}
\usepackage{multirow}
\usepackage{array}
\usepackage{caption}
\usepackage{subcaption}
\usepackage[export]{adjustbox}
\graphicspath{{figures/}}

\usepackage{framed}

\usepackage[numbers,sort&compress]{natbib}

\usepackage{hyperref}
\hypersetup{
    colorlinks=true,
    linkcolor=black,
    citecolor=blue,
    urlcolor=blue,
    pdfauthor={Aman Kumar Pandey and Anil Singh Parihar},
    pdftitle={LUMINA-26: Low-Light Understanding for Modeling and Interpreting Night-time Actions}
}

\colorlet{shadecolor}{yellow}
\definecolor{newcolor}{rgb}{.8,.349,.1}
\begin{document}

\begin{center}
\vspace*{-0.25in}

{\LARGE\bfseries LUMINA-26: Low-Light Understanding for Modeling and Interpreting Night-time Actions\par}

\vspace{0.9em}

{\large
Aman Kumar Pandey$^{1,2}$ \quad
Anil Singh Parihar$^{1}$
\par}

\vspace{0.5em}

{\small
$^{1}$Department of Computer Science and Engineering, Delhi Technological University, Delhi, India
\par}

{\small
$^{2}$Department of Computer Science, Delhi College of Arts and Commerce, University of Delhi, Delhi, India
\par}

\vspace{0.45em}

{\small
\texttt{amanpnd60@outlook.com} \quad
\texttt{parihar.anil@gmail.com}
\par}

\end{center}

\vspace{0.75em}

\begin{abstract}
Low-light human action recognition remains a challenging problem due to poor illumination, amplified noise, motion ambiguity, and diverse real-world scenes. Existing low-light datasets often lack sufficient action diversity, capture realism, or balanced class distribution, limiting the development of robust models. To address this, we introduce \textbf{LUMINA-26}: \textbf{L}ow-Light \textbf{U}nderstanding for \textbf{M}odeling and \textbf{I}nterpreting \textbf{N}ight-time \textbf{A}ctions, comprising 6,784 clips across 26 action classes, recorded from 22 subjects across 20 indoor and outdoor locations under naturally occurring low-light conditions. We also propose Illumi-Net: An Illumination-Adaptive Mixture-of-Experts Network, which leverages video-level illumination cues to guide adaptive enhancement and transformer-based spatio-temporal feature extraction, with expert-conditioned decision fusion. Our method surpasses previous state-of-the-art performance on ELLAR (Top-1: 55.13\%, Top-5: 78.87\%) and establishes a strong baseline on LUMINA-26 (Top-1: 75.95\%, Top-5: 93.58\%), offering a practical benchmark for future low-light action recognition research.
\end{abstract}

\vspace{0.5em}

\vspace{1.2em}

\section{Introduction}

Human Action Recognition (HAR) has become a cornerstone problem in computer vision, driven by applications such as intelligent surveillance, healthcare monitoring, autonomous systems, sports analytics, and human--computer interaction \cite{1,2,3,4,5}. Recent advances in deep learning have significantly improved recognition accuracy under controlled, well-lit environments. In particular, state-of-the-art deep neural architectures based on convolutional and spatio-temporal modeling \cite{wang2018non,tran2018closer,tran2019video,tran2015learning,feichtenhofer2019slowfast,carreira2017quo} and recent video transformers \cite{arnab2021vivit,bertasius2021space,fan2021multiscale,liu2022video} have propelled HAR performance. Further improvements include attention-based feature fusion, residual attention mechanisms, and efficient adapter-based video learning strategies \cite{zhao2025attention,lu2025cache}. However, most existing HAR methods and benchmarks primarily rely on videos captured in normal-light conditions \cite{ucf101,kinetics400,kinetics600,kinetics700,somethingv1v2r4}, limiting their effectiveness in real-world low-light scenarios.
In practical settings such as night-time surveillance, traffic monitoring, or public safety applications, vision-based methods often operate under low-light or poorly illuminated conditions. These environments present challenges including low signal-to-noise ratios, reduced contrast, motion blur, and loss of fine-grained visual cues, which can drastically degrade conventional HAR model performance \cite{6,7}. Consequently, models trained on well-lit datasets often fail to generalize reliably in these conditions.
The performance of deep learning models depends on both architectural representational capacity and the quality and diversity of training data. For low-light HAR, research has focused on: \textbf{(i)} designing robust models capable of extracting discriminative features from degraded inputs, and \textbf{(ii)} creating datasets capturing real-world low-light challenges. Despite architectural enhancements and illumination-aware preprocessing, the lack of suitable datasets and adaptive recognition frameworks remains a major bottleneck.
Existing HAR datasets fall into two categories: \textit{well-lit datasets} and \textit{low-light datasets}. Well-lit datasets such as UCF101, Kinetics-400/600/700, and Something-Something V1/V2 provide high-quality annotations and diverse actions \cite{ucf101,kinetics400,kinetics600,kinetics700,somethingv1v2r4}, but are unsuitable for modeling low-light scenarios due to limited illumination variability.
Low-light HAR datasets can be \textit{synthetic} or \textit{real-world}. Synthetic datasets artificially darken well-lit videos but fail to model real-world effects such as sensor noise, non-uniform illumination, and motion artifacts. Real-world low-light datasets, including ARID \cite{arid}, Dark-48 \cite{dark48}, and ELLAR \cite{ha2024ellar}, provide genuine challenges but are limited in action classes, scene diversity, or environmental coverage.
To address these limitations, we introduce \textbf{LUMINA-26}: \textbf{L}ow-Light \textbf{U}nderstanding for \textbf{M}odeling and \textbf{I}nterpreting \textbf{N}ight-time \textbf{A}ctions, a real-world low-light HAR dataset with 6,784 videos across 26 action classes. LUMINA-26 spans diverse indoor and outdoor night-time environments, capturing authentic low-light conditions without synthetic post-processing. The dataset maximizes contextual, spatial, and illumination diversity, providing a robust benchmark for model evaluation.
Additionally, we propose an Illumi-Net: Illumination-aware action recognition framework for low-light videos. Unlike conventional approaches, our framework adaptively enhances illumination prior to spatio-temporal feature extraction, enabling models to preserve discriminative cues across varying darkness levels. With a VideoMAE backbone, this approach achieves robust feature learning in challenging night-time scenarios.

The main contributions of this work are:
\begin{itemize}
    \item We introduce LUMINA-26, a large-scale, authentic night-time human action recognition dataset with 6,784 videos spanning 26 diverse action classes, captured across 22 actors and 20 indoor/outdoor locations. The dataset features balanced class distributions, broad illumination variability, and statistically validated diversity, offering a robust benchmark for evaluating low-light HAR models under real-world conditions.
    \item We propose Illumi-Net, a novel framework that adaptively modulates video inputs according to estimated illumination conditions using a mixture-of-experts strategy. By integrating an illumination descriptor, adaptive gating, and expert-conditioned classification, the model robustly extracts spatio-temporal features in extreme low-light scenarios, outperforming conventional single-strategy enhancement approaches.
    \item Extensive analysis of LUMINA-26 alongside existing low-light HAR datasets, including ELLAR, ARID, and Dark-48, demonstrating the proposed dataset’s diversity, balance, and practical relevance for low-light action recognition.
\end{itemize}
\begin{figure}[htbp]
\centering
\includegraphics[width=0.95\textwidth]{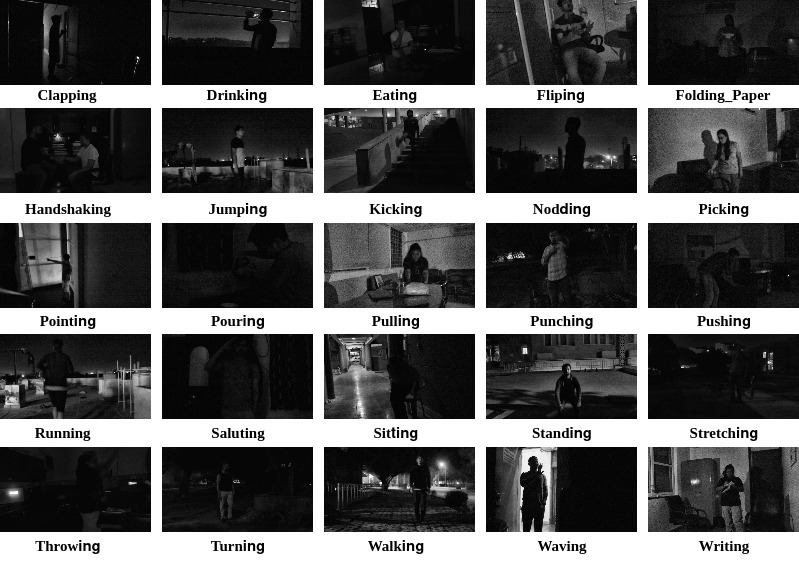}
\caption{Sample frames from LUMINA-26, brightness adjusted for display clarity.}
\label{fig:lumina26_samples}
\end{figure}

The remainder of the paper is structured as follows. Section II reviews related work and existing datasets. Section III introduces the proposed LUMINA-26 dataset. Section IV details the proposed illumination-adaptive recognition framework. Section V describes the experimental setup, while Section VI presents the results and discussion. Section VII concludes the paper, followed by the acknowledgment.

\section{Related Work and Datasets}
\subsection{Action Recognition Datasets}

The development of HAR has been closely intertwined with the availability of benchmark datasets. Early datasets such as KTH, Weizmann, and IXMAS provided controlled environments for evaluating basic actions, but their limited scale and scene diversity restricted robust spatio-temporal modeling. Later, large-scale datasets including HMDB51, UCF101, ActivityNet, Kinetics, and Something-Something introduced richer action vocabularies, greater intra-class variation, and more realistic contexts, which facilitated the advancement of 2D CNNs, 3D CNNs, and transformer-based HAR methods \cite{kthdataset,weizmanndataset,inria,hmdb51,ucf101,activitynet,kinetics400,kinetics600,kinetics700,somethingv1v2r4,aman}.

\begin{table}[htbp]
\centering
\caption{Popular normal-light video datasets for human action recognition.}
\label{datasettable}
\renewcommand{\arraystretch}{1.1}
\footnotesize
\begin{tabular}{cccc}
\toprule
\textbf{Dataset} & \textbf{Year} & \textbf{No. of Videos} & \textbf{No. of Actions} \\
\midrule
KTH \cite{kthmultiview}  & 2004 & 24 & 6 \\
Weizmann \cite{weizmanndataset} & 2005 & 90 & 10 \\
INRIA XMAS \cite{inria} & 2006 & 390 & 13 \\
UCF Sports \cite{Soomro2014} & 2008 & 150 & 10 \\
Hollywood2 \cite{hollywood} & 2009 & 3,669 & 12 \\
UCF11 \cite{ucf11} & 2009 & 1,100+ & 11 \\
HMDB51 \cite{hmdb51}  & 2011 & 6,849 & 51 \\
UCF50 \cite{ucf50} & 2012 & 50 & 50 \\
UCF101 \cite{ucf101} & 2012 & 13,320 & 101 \\
UT Kinect-A \cite{Kinect}  & 2012 & 200 & 10 \\
THUMOS'14 \cite{thumos}  & 2014 & 413 & 20 \\
Sports-1M \cite{sports1m} & 2014 & 1,133,158 & 487 \\
ActivityNet \cite{activitynet}  & 2015 & 28,000 & 203 \\
YouTube-8M \cite{youtube8m}  & 2016 & 8,000,000 & 4,716 \\
NTU-RGB+D \cite{nturgbd} & 2016 & 56,680 & 120 \\
Kinetics \cite{kinetics400, kinetics600, kinetics700}  & 2017 & 500,000 & 600 \\
Something-Something \cite{somethingv1v2r4} & 2017 & 108,499 & 174 \\
AVA \cite{ava} & 2018 & 57,600 & 80 \\
MultiSports \cite{multisports} & 2021 & 3,200 & 4 \\
FineDiving \cite{xu2022finediving} & 2022 & 3,000 & 52 \\
Aeriform In-Action \cite{kapoor2023aeriform} & 2023 & 32 & 13 \\
Ego-Exo4D \cite{grauman2024ego} & 2024 & 5,035 & 43 \\
\bottomrule
\end{tabular}
\end{table}
While these datasets have been instrumental for advancing HAR, they are largely collected under normal illumination, limiting their utility for low-light conditions. Consequently, models trained on these datasets often perform suboptimally in night-time or dimly lit environments, where visibility, motion boundaries, and contextual cues are degraded.

\subsection{Low-Light Video Datasets}

Recent research has introduced datasets for low-light image enhancement and restoration, such as LOL, SID, and ExDark \cite{wei2018deep,sid,exlpose}, but these lack temporal structure for action recognition. To address low-light HAR, dedicated video datasets have emerged:

\begin{itemize}
    \item \textbf{ARID} \cite{arid} is among the first low-light HAR datasets, demonstrating the challenges of real dark videos. Its small number of actions and center-biased compositions can encourage models to exploit positional priors.
    \item \textbf{Dark-48} \cite{dark48} increases scale and semantic diversity by selecting dark clips from existing datasets. While it improves class coverage, it does not fully capture real-world low-light complexities like sensor noise and non-uniform illumination.
    \item \textbf{ELLAR} \cite{ha2024ellar} focuses on extremely low-light scenes with severe visibility degradation. Though valuable, its action vocabulary is limited, and it emphasizes extremely dark conditions rather than general night-time variability.
\end{itemize}
These datasets reveal a clear evolution: ARID introduces the problem, Dark-48 expands scale and diversity, and ELLAR pushes extreme illumination. Nevertheless, each addresses only part of the challenge, highlighting the need for more realistic and diverse low-light benchmarks.

\begin{table*}[htbp]
\centering
\caption[Comparison of representative low-light HAR datasets]{Comparison of representative low-light HAR datasets. UCI denotes Unique Capture Instances, CDUS denotes Composite Dataset Utility Score, and class balance is measured using Shannon evenness.}
\label{tab:lowlight_dataset_comparison}
\renewcommand{\arraystretch}{1.1}
\setlength{\tabcolsep}{3.8pt}
\footnotesize
\begin{tabular}{cccccccccc}
\toprule
\textbf{Dataset} & \textbf{Year} & \textbf{Capture Type} & \textbf{Clips} & \textbf{UCI} & \textbf{Actions} & \textbf{Subjects} & \textbf{Locations} & \textbf{CDUS} & \textbf{Class Balance} \\
\midrule
ARID \cite{arid}& 2021 & Realistic low-light benchmark & 3,780 & 3,780 & 11 & 11 & 18 & 0.448 & 0.990 \\
\textit{*}Dark-48 \cite{dark48} & 2023 & Darkness-filtered video collection & 8,815 & - & 43 & - & - & 0.44 & 0.945 \\
ELLAR \cite{ha2024ellar} & 2024 & Real-world extreme low-light & 12,078 & 4,026 & 12 & 5 & 5 & 0.508 & 0.996 \\
\textbf{LUMINA-26} & \textbf{2026} & \textbf{Real-world low-light} & \textbf{6,784} & \textbf{6,784} & \textbf{26} & \textbf{22} & \textbf{20} & \textbf{0.617} & \textbf{0.997} \\
\bottomrule
\end{tabular}

\vspace{1pt}
\raggedright
\scriptsize \textit{*}Dark-48 is a derived dataset curated from existing video datasets; therefore, UCI, subject count, and location count are not explicitly reported.
\end{table*}

\subsection{Low-Light Action Recognition Methods}

Existing low-light HAR methods generally follow two strategies: (i) preprocessing videos via enhancement or (ii) joint learning of enhancement and recognition. Traditional techniques include histogram equalization \cite{abdullah2007dynamic,celik2011contextual,trahanias1992color}, gamma correction \cite{huang2012efficient,poynton2012digital}, Retinex-based methods \cite{li2018structure,rahman1996multi}, and learning-based enhancement approaches \cite{guo2020zero,guo2016lime,wei2020physics,wu2022uretinex,zhang2019kindling}. Recent enhancement methods based on contrast correction, illumination estimation, deep simultaneous estimation, and quaternion representations have further improved visual details under low-light conditions \cite{parihar2017fuzzy,singh2023dse,singh2024illumination,acharjee2025qlight}. While these methods improve perceptual quality, they are primarily designed for image-level enhancement and may not preserve discriminative motion cues in video-based recognition.
Adaptive low-light HAR models such as DarkLight and illumination-guided frameworks improve recognition by incorporating illumination information \cite{darklight,dark48}. Limitations persist, including fixed preprocessing, multi-stage pipelines, and insufficient preservation of spatio-temporal cues across varying darkness. 

\subsection{Limitations of Existing Low-Light HAR Benchmarks and Methods}
Despite recent progress, challenges remain. Existing datasets are constrained by limited actions, scene diversity, or construction from filtered clips rather than real low-light acquisition. Methods often rely on sample-independent enhancement or expensive multi-stage designs, which may degrade spatio-temporal features. These gaps motivate both the creation of LUMINA-26 and the development of illumination-aware recognition frameworks for real-world night-time HAR.

\section{The Proposed LUMINA-26 Dataset}
To address the limitations of current low-light human action recognition datasets, we present \textbf{LUMINA-26}: \textbf{L}ow-Light \textbf{U}nderstanding for \textbf{M}odeling and \textbf{I}nterpreting \textbf{N}ight-time \textbf{A}ctions, a real-world low-light video benchmark designed for robust night-time action recognition. LUMINA-26 emphasizes authentic acquisition, diverse and semantically meaningful action classes, environmental variability, and class balance. Unlike datasets generated by synthetic darkening or filtered from pre-existing video corpora, LUMINA-26 is directly recorded under naturally dim conditions, preserving realistic degradations such as low contrast, sensor noise, shadows, and motion blur. This design allows evaluation of not only accuracy but also model robustness and interpretability under practical low-light scenarios.

\subsection{Data Acquisition and Pipeline}
Videos were captured using high-resolution DSLR cameras (Canon 1300D, Canon 200D, Nikon B700) mounted on tripods to ensure stability and minimize motion artifacts. Recordings were conducted at $1920 \times 1080$ resolution and 25 fps, spanning over 9 indoor and 11 outdoor locations across Delhi. A total of 22 volunteers (15 male, 7 female) contributed, providing diversity in body shapes, postures, motion styles, and interactions.

Raw videos underwent the following processing:
\begin{itemize}
    \item Conversion from \texttt{.MOV} to \texttt{.mp4} format,
    \item Manual verification and sorting into correct action classes,
    \item Removal of audio tracks to maintain privacy,
    \item Exclusion of clips shorter than 2 s to ensure temporal consistency.
\end{itemize}

This pipeline produces a dataset that is visually authentic while remaining methodologically robust for reproducible spatio-temporal modeling.

\subsection{Dataset Composition and Action Taxonomy}

LUMINA-26 contains 6,784 clips across 26 action classes, covering both single-person motions (with/without objects) and multi-person interactions. Each class contains roughly 260 videos, resulting in a balanced distribution (coefficient of variation = 0.216, Shannon evenness = 0.992). Fig.~\ref{fig:class_count_comparison} shows that LUMINA-26 maintains a balanced class-wise distribution across the training, validation, and test splits. Most classes contain roughly similar numbers of clips, with only modest variation across categories. This balanced structure improves the benchmark quality of the dataset by reducing class dominance, supporting more stable optimization during training, and enabling a fairer assessment of model generalization. Clip durations are approximately 2 seconds. Evaluation protocols include random splits, cross-subject, and cross-location schemes.
\begin{table*}[htbp]
\captionsetup{justification=raggedright,singlelinecheck=false} 
\caption[Summary statistics of LUMINA-26]{Summary statistics of LUMINA-26. Darkness-related statistics are derived from the per-video brightness analysis.}
\label{tab:lumina26_statistics}
\renewcommand{\arraystretch}{1.15}
\setlength{\tabcolsep}{6pt}
\centering
\begin{tabular}{cp{5.2cm}p{6.8cm}}
\toprule
\textbf{Category} & \textbf{Characteristics} & \textbf{Value} \\
\midrule
\multirow{6}{*}{\textbf{Composition}}
& Number of action classes & 26 \\
& Total number of clips & 6784 \\
& Total duration & 13,550 s ($\approx$ 3.76 h) \\
& Average clip duration & $\approx$ 2 s \\
& Number of actors & 22 \\
& Number of recording locations & 20 \\ \midrule
\multirow{5}{*}{\textbf{Acquisition}}
& Capture condition & Natural low-light / night-time \\
& Frame rate & 25 fps \\
& Spatial resolution & 1920 $\times$ 1080 \\
& Camera types & Canon 1300D, Canon 200D, Nikon B700 \\
& Audio track & Removed \\ \midrule
\multirow{4}{*}{\textbf{Class Statistics}}
& Class count range & 91 -- 325 \\
& Average videos per class & 260.9 \\
& Class count coefficient of variation & 0.216 \\
& Class balance (Shannon evenness) & 0.992 \\ \midrule
\multirow{4}{*}{\textbf{Illumination}}
& Mean darkness index & 99.14 \\
& Median darkness index & 99.67 \\
& Minimum darkness index & 85.16 \\
& Maximum darkness index & 99.99 \\ \midrule
\multirow{1}{*}{\textbf{Protocols}}
& Supported evaluation protocols & Random split, cross-subject, cross-location \\
\bottomrule
\end{tabular}
\end{table*}
\subsection{Comparative Benchmark Analysis}
\begin{figure}[htbp]
\centering
\includegraphics[width=0.95\textwidth]{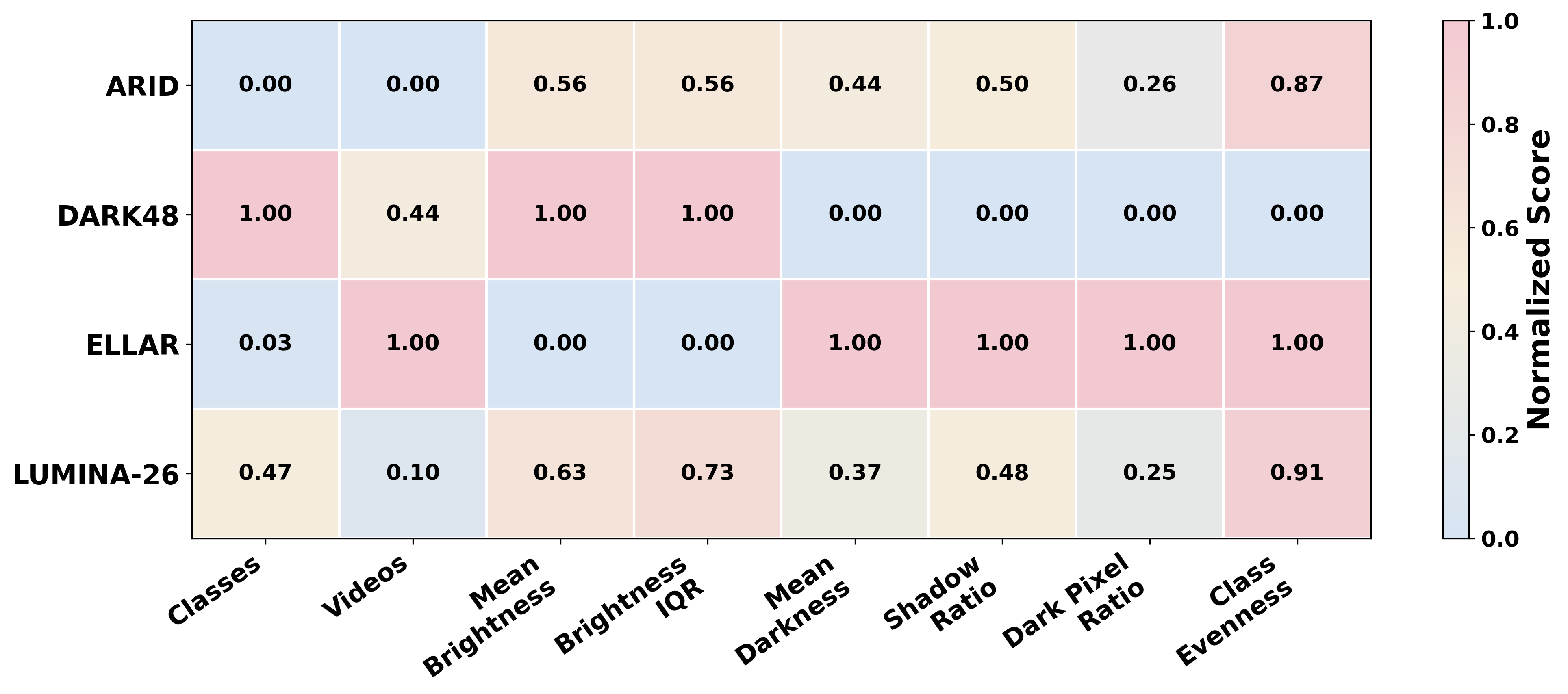}
\caption{\footnotesize Normalized multi-metric comparison of low-light HAR datasets. LUMINA-26 demonstrates balanced diversity, illumination challenge, and class representation.}
\label{fig:dataset_metric_heatmap}
\end{figure}

\begin{figure}[t]
\centering
\includegraphics[width=0.98\textwidth]{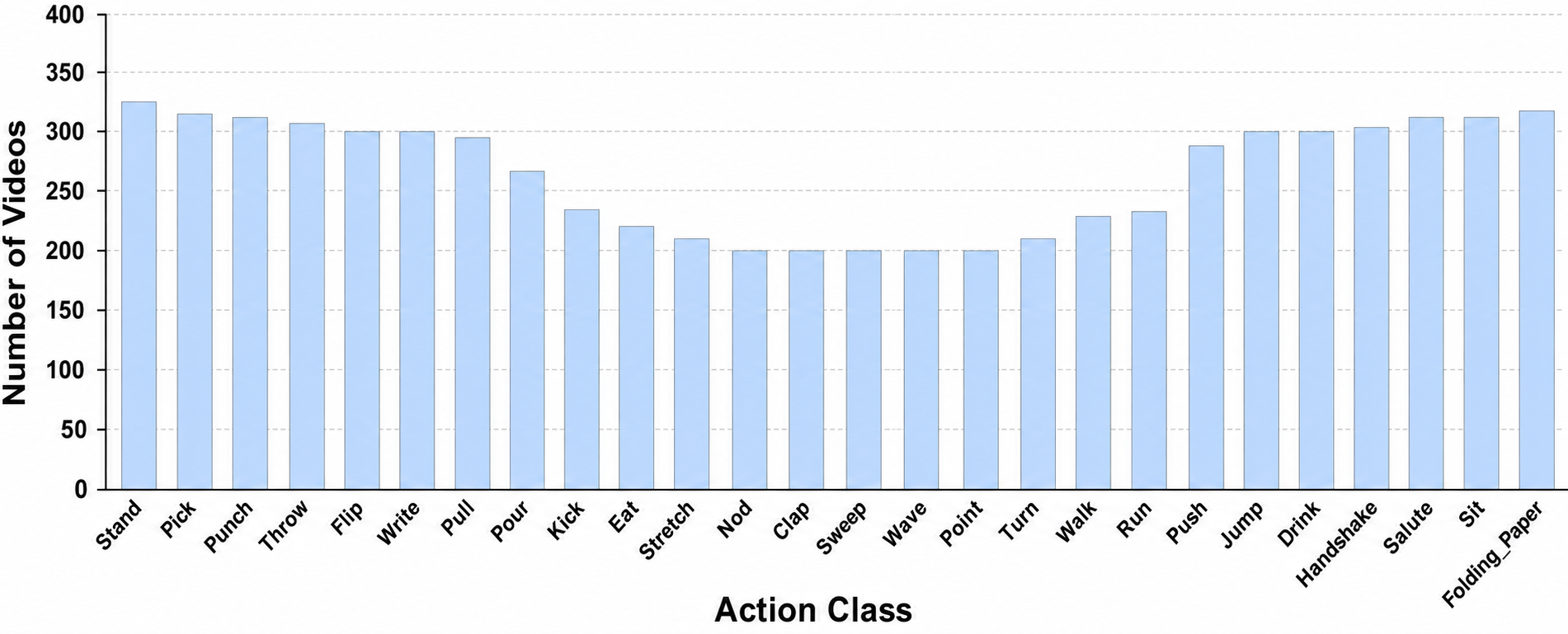}
\caption{Class-wise distribution of LUMINA-26 across the training, validation, and test splits. Most categories contain 200–325 clips, supporting balanced training and evaluation.}
\label{fig:class_count_comparison}
\end{figure}

\begin{figure}[t]
\centering
\includegraphics[width=0.95\textwidth]{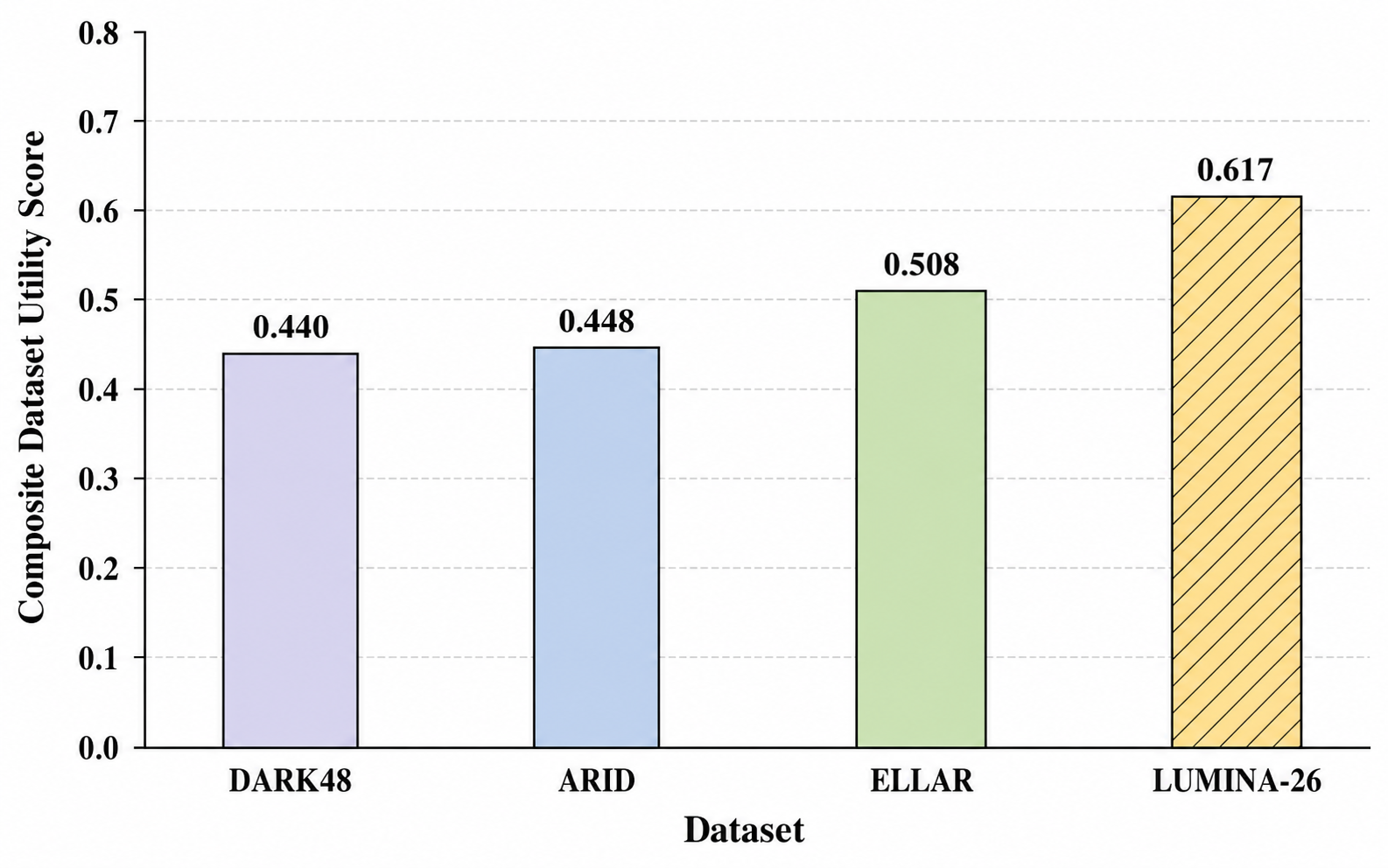}
\caption{Composite dataset utility score (CDUS) among low-light HAR datasets. LUMINA-26 achieves the highest score, indicating superior benchmark quality.}
\label{fig:dataset_quality_scorecard}
\end{figure}

Figure \ref{fig:dataset_metric_heatmap} presents a normalized multi-metric comparison of LUMINA-26 with ARID, Dark-48, and ELLAR, encompassing class diversity, dataset size, brightness distribution, darkness severity, shadow ratio, and class evenness. LUMINA-26 demonstrates a well-balanced profile across all these metrics. The class-wise distribution in Figure \ref{fig:class_count_comparison} shows uniform representation across categories, while the composite dataset utility score (CDUS) illustrated in Figure \ref{fig:dataset_quality_scorecard}, summarizes the overall benchmark quality by integrating action diversity, dataset size, class balance, darkness, and shadow severity. LUMINA-26 achieves the highest score, emphasizing its superior balance between low-light challenge, semantic coverage, and benchmark usability.

\subsection{Low-Light Statistical Characterization}

\begin{figure}[t]
\centering
\includegraphics[width=0.95\textwidth]{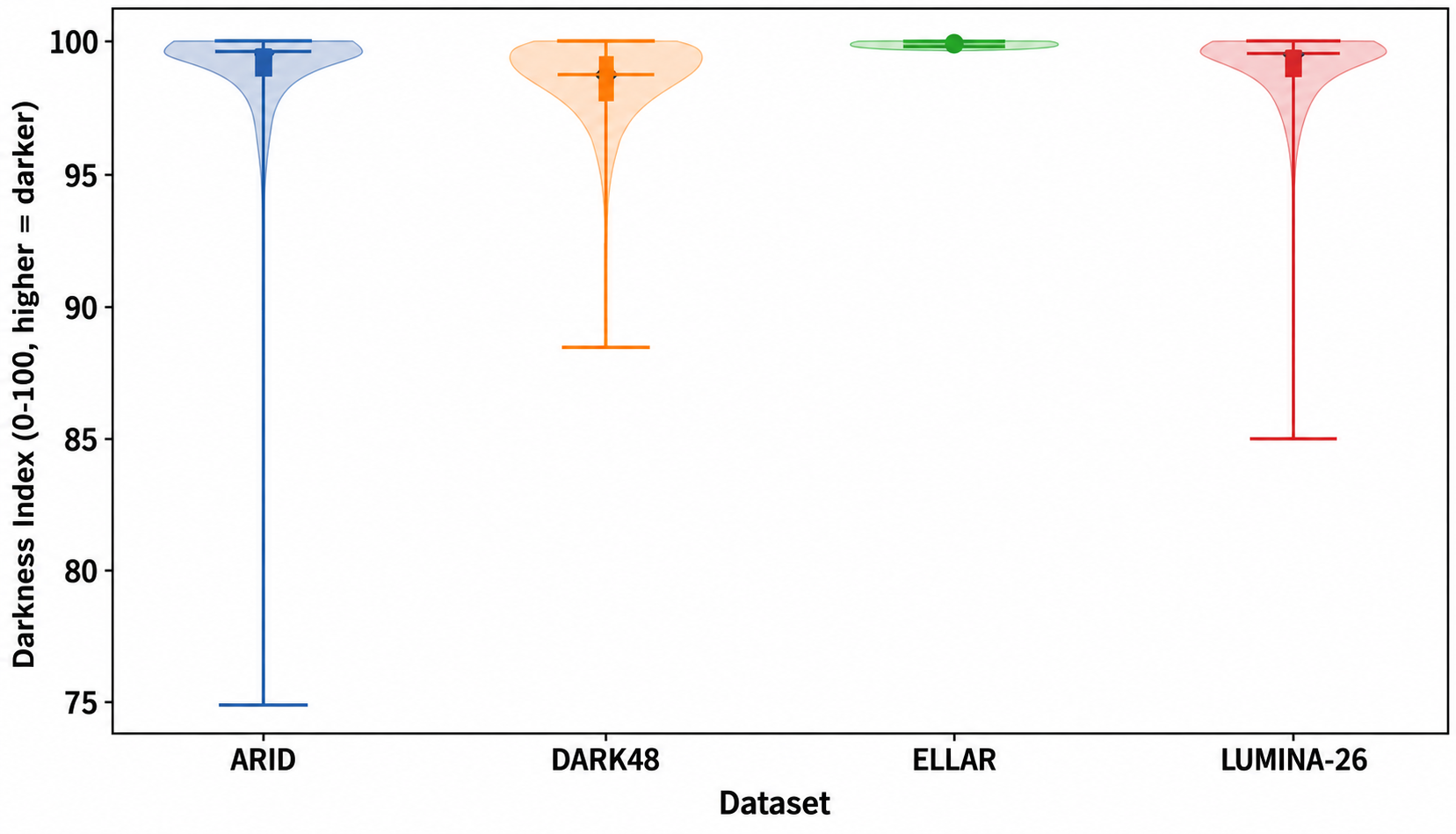}
\caption{Darkness distribution comparison across low-light HAR datasets. LUMINA-26 provides strong low-light conditions while avoiding extreme saturation.}
\label{fig:violin_darkness}
\end{figure}

\begin{figure}[t]
\centering
\includegraphics[width=0.95\textwidth]{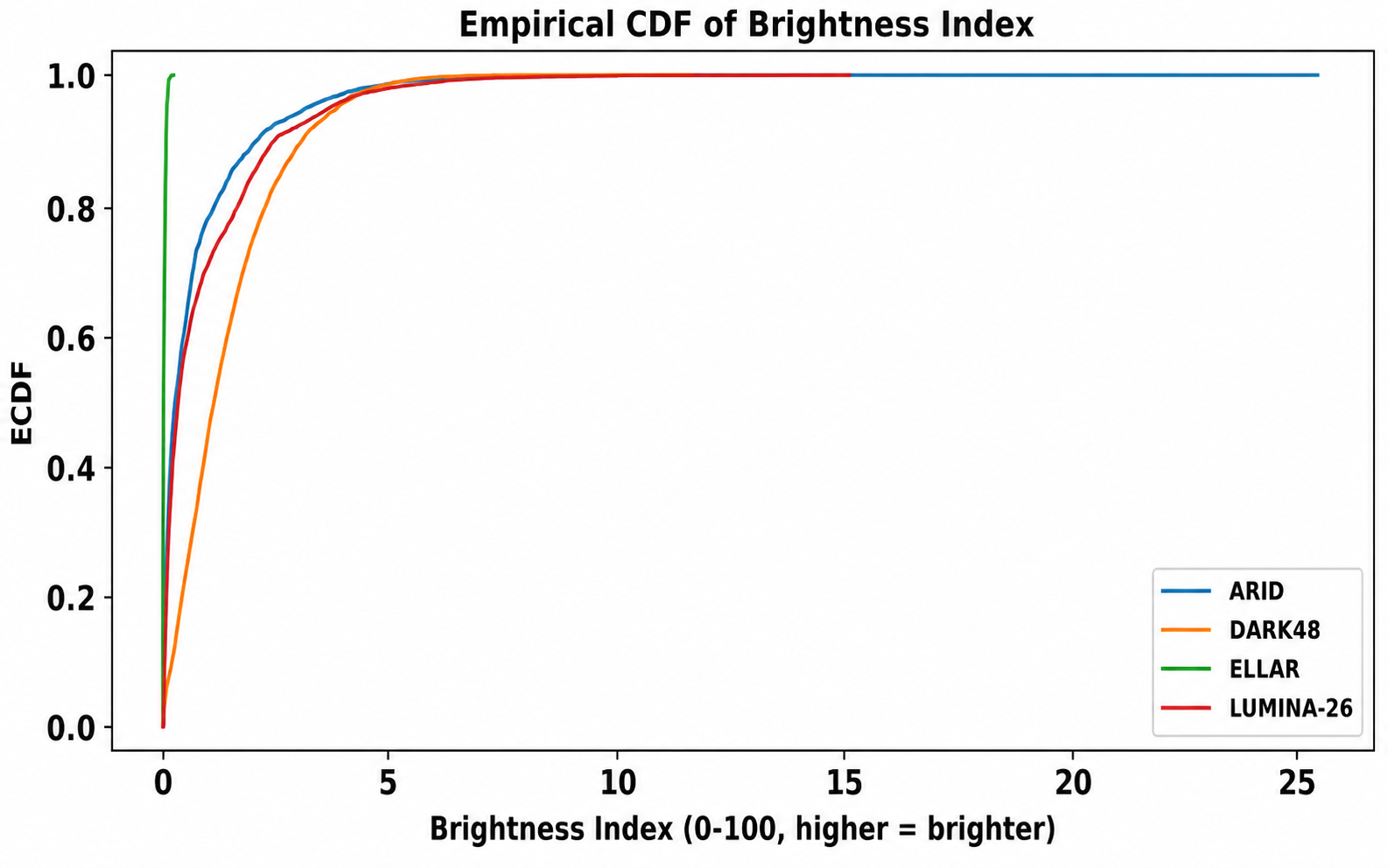}
\caption{Empirical cumulative distribution of brightness index across low-light HAR datasets. LUMINA-26 follows low-brightness tendencies without collapsing to extreme darkness.}
\label{fig:ecdf_brightness}
\end{figure}

The frame-level luminance is calculated using the standard Rec. 601 formula:

\begin{equation}
Y = 0.299R + 0.587G + 0.114B,
\end{equation}

where $R$, $G$, and $B$ are the red, green, and blue channel intensities, respectively. The brightness index (BI) of a frame is computed as the mean luminance over all pixels:

\begin{equation}
BI = \frac{1}{H W} \sum_{x=1}^{H} \sum_{y=1}^{W} Y(x,y),
\end{equation}

where $H$ and $W$ denote the frame height and width. The darkness index (DI) is defined as:

\begin{equation}
DI = 100 \times (1 - BI).
\end{equation}
For video-level analysis, the BI and DI are averaged across all sampled frames in a clip. Figures \ref{fig:violin_darkness} and \ref{fig:ecdf_brightness} depict the distributions and empirical cumulative distribution functions of darkness/brightness, showing that LUMINA-26 captures authentic low-light conditions while avoiding the extreme darkness observed in ELLAR. Statistical significance is confirmed through Kruskal--Wallis and pairwise Mann--Whitney tests, indicating that LUMINA-26 is distinct from ARID, Dark-48, and ELLAR.
\subsection{Overall Benchmark Utility}
LUMINA-26 strikes a favorable balance between realism, low-light severity, semantic diversity, and class distribution. While ELLAR captures extreme darkness and Dark-48 emphasizes broader class coverage, LUMINA-26 provides authentic low-light acquisition with a richer action vocabulary than ARID and ELLAR, alongside substantially improved class balance relative to Dark-48. CDUS integrating diversity, size, balance, darkness, and shadow severity further confirms LUMINA-26 as a balanced and practically relevant benchmark.

\section{Proposed Approach}
\label{sec:proposed}
HAR in extreme low-light environments remains challenging because insufficient illumination causes low contrast, amplified sensor noise, loss of fine spatial details, and degraded temporal consistency, all of which hinder robust spatio-temporal representation learning. Most existing approaches process input videos using a fixed pipeline, implicitly assuming that visual quality remains relatively consistent across samples. In real-world low-light scenarios, however, illumination can vary substantially not only across datasets but also within individual video clips. As a result, a uniform processing strategy is inherently suboptimal.
To address this limitation, we propose an Illumi-Net: An Illumination-Adaptive Mixture-of-Experts Network that explicitly models the lighting condition of each video and uses this information to guide both enhancement and classification. The key idea is to first estimate the degree of darkness or degradation in a video, then adaptively enhance it using a set of specialized expert transformations, and finally perform classification using an illumination-aware decision mechanism.
\begin{figure*}[!t]
\centering
\includegraphics[width=0.93\textwidth,height=0.41\textheight]{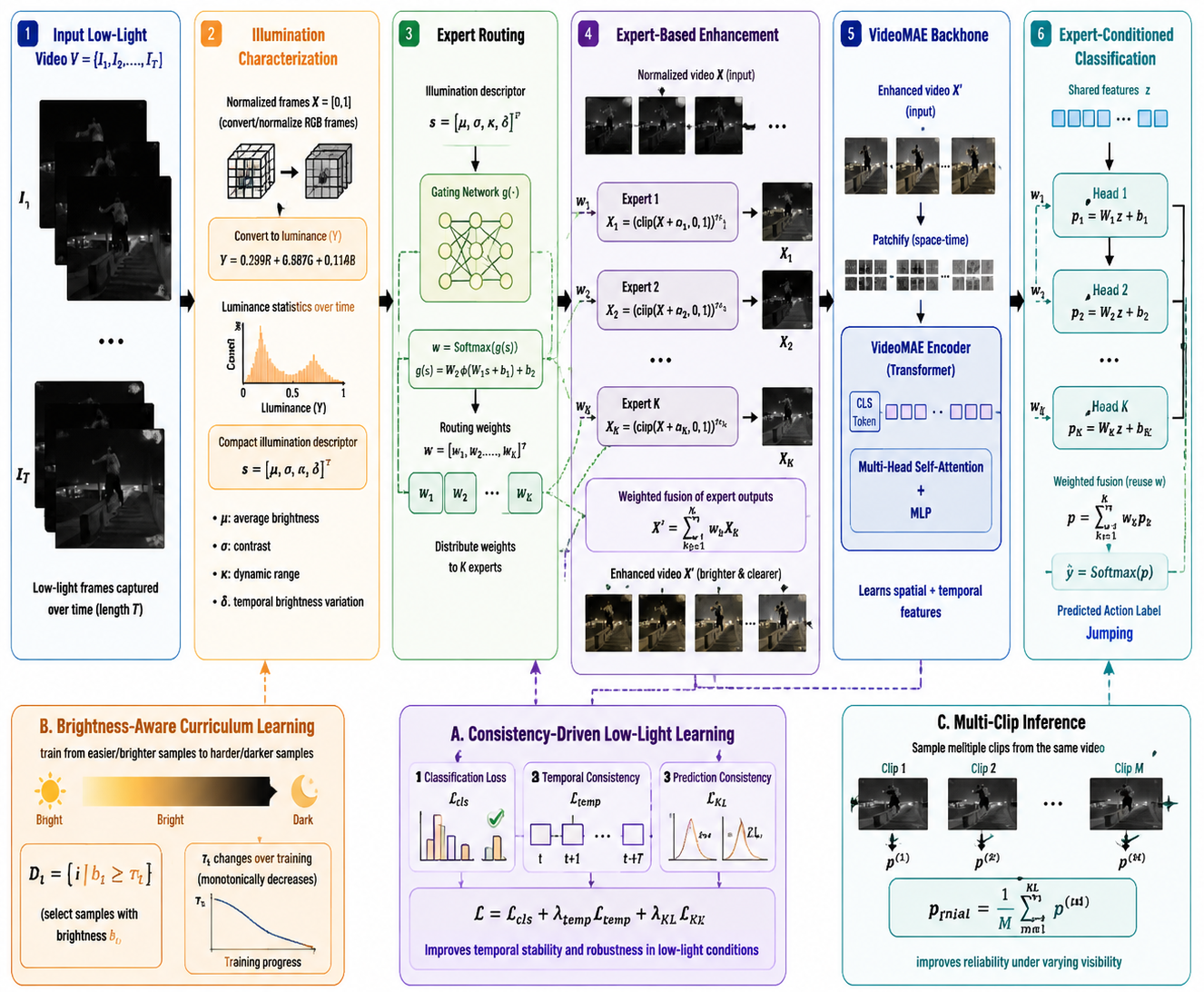}
\caption{Illustration of the proposed Illumi-Net: An Illumination-Adaptive Mixture-of-Experts Network for low-light HAR. An illumination descriptor guides a gating network to assign weights to multiple enhancement experts. The fused output is processed by a VideoMAE backbone for spatio-temporal feature extraction, followed by expert-conditioned classification. The model is trained with classification, temporal consistency, and prediction consistency losses, and uses brightness-aware curriculum learning. Final predictions are obtained via multi-clip inference for improved robustness under varying low-light conditions.}
\label{fig:proposed_framework}
\end{figure*}
The proposed framework consists of four main stages:
\begin{itemize}
    \item \textbf{Illumination characterization}, where a compact descriptor is extracted to summarize the lighting condition of each video.
    \item \textbf{Adaptive enhancement via mixture-of-experts}, where multiple expert transformations are selectively combined according to the illumination profile.
    \item \textbf{Spatio-temporal feature extraction and classification}, where the enhanced video is processed by a transformer backbone and classified using expert-conditioned fusion.
    \item \textbf{Consistency-driven learning}, which promotes temporal stability and robustness under perturbations.
\end{itemize}
In addition, a brightness-aware curriculum learning strategy is employed to stabilize training and progressively expose the model to more challenging samples.
Let a video clip be represented as
\begin{equation}
\mathcal{V} = \{ I_1, I_2, \dots, I_T \},
\label{eq:video}
\end{equation}
where $I_t \in \mathbb{R}^{H \times W \times 3}$ denotes the $t$-th RGB frame and $T$ is the total number of frames.
The goal of HAR is to learn a mapping
\begin{equation}
f: \mathcal{V} \rightarrow y,
\label{eq:mapping}
\end{equation}
where $y \in \{1,2,\dots,C\}$ is the action label.

In standard settings, this mapping assumes that both spatial features, such as object shape and appearance, and temporal features, such as motion patterns, are clearly observable. Under low-light conditions, however, this assumption no longer holds. Important regions may become indistinguishable, edges may blur, and motion cues may be obscured by unstable brightness.
To explicitly incorporate illumination awareness, we reformulate the mapping as
\begin{equation}
f(\mathcal{V}) = h\big(\Phi(\Psi(\mathcal{V}; \mathbf{s})); \mathbf{s}\big),
\label{eq:framework}
\end{equation}
where $\mathbf{s}$ is an illumination descriptor extracted from the video. In this formulation, $\Psi(\cdot)$ denotes the enhancement function that adapts the video according to illumination, $\Phi(\cdot)$ represents the feature extraction backbone, and $h(\cdot)$ denotes the classification function. This decomposition highlights that both representation learning and classification are conditioned on the illumination characteristics of the input, making the overall framework more adaptive and robust.

\subsubsection{Illumination Characterization}
\label{subsubsec:illumination}
Given normalized frames $X \in [0,1]^{T \times H \times W \times 3}$, we first compute luminance as
\begin{equation}
Y = 0.299R + 0.587G + 0.114B.
\label{eq:luminance}
\end{equation}
This transformation converts RGB values into a single intensity channel that reflects perceived brightness, enabling the model to focus on illumination while remaining less sensitive to color variations.
To summarize the lighting condition of the entire video, we compute four statistical descriptors:
\begin{align}
\mu &= \mathbb{E}[Y], \label{eq:mean}\\
\sigma &= \sqrt{\mathbb{E}[(Y - \mu)^2]}, \label{eq:std}\\
\kappa &= P_{95}(Y) - P_{5}(Y), \label{eq:range}\\
\delta &= \mathbb{E}[|Y_{t+1} - Y_t|]. \label{eq:temporal}
\end{align}
Each descriptor has a specific interpretation:
\begin{itemize}
    \item $\mu$ denotes the average brightness of the video, where a lower value indicates a darker scene.
    \item $\sigma$ measures the spread of intensity values and reflects contrast. Low contrast generally implies reduced visibility of structures.
    \item $\kappa$ captures the effective dynamic range, indicating the extent of variation between bright and dark regions.
    \item $\delta$ measures temporal changes in brightness and helps identify flickering or unstable illumination.
\end{itemize}
These descriptors are combined into the vector
\begin{equation}
\mathbf{s} = [\mu, \sigma, \kappa, \delta],
\label{eq:descriptor}
\end{equation}
which serves as a compact representation of the illumination condition of the video.

\subsubsection{Expert Routing Mechanism}
\label{subsubsec:routing}
Given the illumination descriptor $\mathbf{s}$, the model determines how strongly each expert should contribute to the enhancement process. This is achieved through a gating function:
\begin{equation}
\mathbf{w} = \mathrm{Softmax}(g(\mathbf{s})),
\label{eq:weights}
\end{equation}
where the Softmax operation ensures that the weights are positive and sum to one, making them interpretable as contribution probabilities.
The gating function is defined as
\begin{equation}
g(\mathbf{s}) = \mathbf{W}_2 \, \phi(\mathbf{W}_1 \mathbf{s} + \mathbf{b}_1) + \mathbf{b}_2,
\label{eq:gating}
\end{equation}
where $\mathbf{W}_1$ and $\mathbf{W}_2$ are learnable weight matrices, $\mathbf{b}_1$ and $\mathbf{b}_2$ are learnable biases, and $\phi(\cdot)$ is a non-linear activation function.
Intuitively, this network learns to map illumination conditions to suitable enhancement strategies. For example, extremely dark videos may activate experts that apply stronger brightness correction, whereas moderately dark videos may favor milder transformations.

\subsubsection{Expert-Based Enhancement}
\label{subsubsec:experts}
Each expert applies a distinct transformation to the input video:
\begin{equation}
X_k = \left( \mathrm{clip}(X + \alpha_k, 0, 1) \right)^{\gamma_k},
\label{eq:expert}
\end{equation}
where $\alpha_k$ controls brightness adjustment and $\gamma_k$ controls contrast through a non-linear transformation.
The clipping operation keeps pixel values within the valid range. Exponentiation by $\gamma_k$ redistributes intensity values, enabling enhancement of dark regions while preserving brighter areas.
The final enhanced video is computed as
\begin{equation}
X' = \sum_{k=1}^{K} w_k X_k.
\label{eq:enhanced}
\end{equation}
This weighted combination allows the model to blend multiple enhancement strategies rather than depending on a single fixed transformation. As a result, the enhancement process becomes adaptive and tailored to the characteristics of each input video.
\subsection{Spatio-Temporal Feature Extraction}
\label{subsec:features}
After illumination-adaptive enhancement, the transformed video $X'$ is passed to a transformer-based backbone for feature extraction. In this work, we adopt VideoMAE as the feature encoder because of its strong ability to capture long-range temporal dependencies and fine-grained motion patterns.
Formally, the feature extraction process is defined as
\begin{equation}
\mathbf{z} = \Phi(X'),
\label{eq:features}
\end{equation}
where $\Phi(\cdot)$ denotes the backbone network and $\mathbf{z} \in \mathbb{R}^{D}$ is the resulting video-level embedding.
The representation $\mathbf{z}$ encodes both spatial and temporal information, including appearance, motion dynamics, and contextual dependencies across frames. It is important to note that the quality of input frames directly affects the quality of the learned representation. In low-light conditions, raw inputs often contain weak or noisy signals, which makes meaningful feature extraction difficult. By introducing illumination-adaptive enhancement before feature extraction, the backbone receives visually improved inputs in which important structures and motion cues are more distinguishable.
Accordingly, the enhancement module and the backbone are not independent components. Rather, they operate jointly: the enhancement stage prepares a representation-friendly input, while the backbone extracts semantically meaningful features from it. This interaction is essential for robust performance in challenging low-light conditions.

\subsection{Expert-Conditioned Classification}
\label{subsec:classification}
Although adaptive enhancement improves input quality, illumination variability may still affect how classification decisions should be made. The appearance of the same action can differ considerably between well-lit and poorly lit conditions. To address this, we extend the mixture-of-experts formulation to the classification stage.
Instead of using a single classifier, we employ $K$ expert-specific classifiers. Each classifier specializes in a different feature distribution associated with a particular range of illumination conditions.
Given the feature representation $\mathbf{z}$ obtained from Eq.~(\ref{eq:features}), the logits produced by the $k$-th expert classifier are
\begin{equation}
\mathbf{p}_k = \mathbf{W}_k \mathbf{z} + \mathbf{b}_k,
\label{eq:logits}
\end{equation}
where $\mathbf{W}_k$ and $\mathbf{b}_k$ are learnable parameters.
The final prediction is then computed as a weighted combination of expert outputs:
\begin{equation}
\mathbf{p} = \sum_{k=1}^{K} w_k \mathbf{p}_k,
\label{eq:fusion}
\end{equation}
where $\mathbf{w}$ is the same routing weight vector obtained from Eq.~(\ref{eq:weights}).
Finally, class probabilities are computed as
\begin{equation}
\hat{\mathbf{y}} = \mathrm{Softmax}(\mathbf{p}).
\label{eq:softmax}
\end{equation}
This formulation ensures that classification remains aligned with the illumination condition of the input. For example, if a video is extremely dark, the routing mechanism can assign higher weights to classifiers that are better suited for low-visibility features.
By coupling enhancement and classification through the same routing weights, the model maintains consistency between how the input is processed and how the final decision is made. This reduces mismatch between representation and classification and leads to more reliable predictions.

\subsection{Consistency-Driven Low-Light Learning}
\label{subsec:loss}
To further improve robustness in extreme low-light conditions, we introduce additional training constraints beyond standard classification supervision. These constraints are designed to preserve temporal structure and maintain prediction stability under photometric perturbations.
\subsubsection{Classification Loss}
The primary objective is the categorical cross-entropy loss:
\begin{equation}
\mathcal{L}_{cls} = - \sum_{c=1}^{C} y_c \log \hat{y}_c,
\label{eq:cls}
\end{equation}
where $y_c$ is the ground-truth label and $\hat{y}_c$ is the predicted probability.
This loss ensures that the model learns discriminative representations for action classification.
\subsubsection{Temporal Consistency Loss}
Although enhancement improves visibility, it may also introduce undesirable temporal artifacts, such as flickering or inconsistent brightness changes between adjacent frames. To reduce such effects, we enforce temporal consistency between the original and enhanced videos:
\begin{equation}
\mathcal{L}_{temp} =
\sum_{t=1}^{T-1}
\left\|
(X'_{t+1} - X'_t) - (X_{t+1} - X_t)
\right\|_2^2.
\label{eq:temp}
\end{equation}
This loss compares frame-to-frame differences before and after enhancement. Ideally, the enhancement process should improve visibility while preserving the underlying motion. If artificial temporal fluctuations are introduced, this loss penalizes such behavior.

\subsubsection{Prediction Consistency Loss}
To improve robustness against photometric perturbations, we generate an augmented version of the enhanced video, denoted by $\tilde{X}$, and compute its prediction $\tilde{\hat{\mathbf{y}}}$. We then enforce consistency between the original and perturbed predictions:
\begin{equation}
\mathcal{L}_{KL} =
D_{KL}(\hat{\mathbf{y}} \parallel \tilde{\hat{\mathbf{y}}}),
\label{eq:kl}
\end{equation}
where $D_{KL}$ denotes the Kullback--Leibler divergence.
This term encourages the model to produce stable predictions even when the input undergoes small photometric changes. Such stability is especially important in low-light settings, where slight illumination fluctuations may otherwise cause large changes in model output.

\subsubsection{Overall Loss Function}
The final training objective is a weighted combination of all losses:
\begin{equation}
\mathcal{L} =
\mathcal{L}_{cls}
+ \lambda_{temp} \mathcal{L}_{temp}
+ \lambda_{KL} \mathcal{L}_{KL},
\label{eq:total}
\end{equation}
where $\lambda_{temp}$ and $\lambda_{KL}$ are balancing hyperparameters.
This objective ensures that the model learns not only to classify correctly, but also to preserve temporal coherence and maintain prediction stability under challenging illumination conditions.

\subsection{Brightness-Aware Curriculum Learning}
\label{subsec:curriculum}
Training directly on heavily degraded low-light samples can destabilize optimization, especially during the early stages when the model has not yet learned robust representations. To alleviate this issue, we adopt a brightness-aware curriculum learning strategy.
Let $b_i$ denote a brightness score associated with sample $i$. At training step $t$, the active training subset is defined as
\begin{equation}
\mathcal{D}_t = \{ i \mid b_i \geq \tau_t \},
\label{eq:dataset}
\end{equation}
where $\tau_t$ is a threshold that evolves over time.
The threshold is defined as
\begin{equation}
\tau_t = \tau_{min} + (\tau_{max} - \tau_{min}) \cdot \frac{t}{T_{train}},
\label{eq:threshold}
\end{equation}
where $T_{train}$ is the total number of training iterations.
At the beginning of training, $\tau_t$ is relatively high, which means the model focuses on brighter and easier samples. As training progresses, the threshold decreases, gradually introducing darker and more difficult samples. This strategy allows the model to first learn stable semantic representations before confronting severely degraded inputs, thereby improving convergence stability and overall performance.

\subsection{Inference Strategy}
\label{subsec:inference}
During inference, relying on a single clip may be insufficient, especially when visibility varies across different parts of the same video. To improve robustness, we adopt a multi-clip evaluation strategy.
Given $M$ clips sampled from a video, predictions are computed for each clip and then aggregated as
\begin{equation}
\mathbf{p}_{final} = \frac{1}{M} \sum_{m=1}^{M} \mathbf{p}^{(m)},
\label{eq:multi}
\end{equation}
where $\mathbf{p}^{(m)}$ is the prediction for the $m$-th clip.
The final class label is then obtained as
\begin{equation}
\hat{y} = \arg\max_c \mathbf{p}_{final}^{(c)}.
\label{eq:final}
\end{equation}
This strategy reduces prediction variance by incorporating evidence from multiple temporal segments of the video. In low-light settings, where some frames may be substantially more informative than others, such aggregation improves prediction reliability.

\section{Experiments}
\label{sec:experiments}
We evaluate the proposed Illumi-Net framework through benchmark comparison and ablation analysis. We first assess the model on the ELLAR benchmark, where it is compared against previously reported low-light HAR methods, and then report its performance on the proposed LUMINA-26 dataset. Together, these experiments examine both cross-benchmark generalization and the practical utility of the proposed design under real low-light conditions.

\subsection{Experimental Setup}
The proposed model is built upon the videomae backbone and is extended with an illumination-aware mixture-of-experts module and expert-conditioned classification heads. For all experiments, we use 16-frame input clips with a temporal sampling rate of 2. Training is performed for 80 epochs using AdamW with an initial learning rate of $3\times10^{-5}$, weight decay of 0.05, and a cosine learning-rate schedule with a warmup ratio of 0.1. In addition, we employ brightness-aware curriculum sampling, DarAug-based low-light augmentation, temporal consistency regularization, and KL-based prediction consistency during optimization.
For validation, we use 3-clip inference, while final test results are reported using 5-clip inference. All reported accuracies are computed at the video level after multi-clip aggregation. For fair evaluation, the final test performance is obtained from the checkpoint with the best validation performance.

\subsection{Evaluation Metrics}

We report Top-1 and Top-5 classification accuracy as the primary evaluation metrics. Top-1 accuracy measures whether the highest-confidence prediction matches the ground-truth action label, while Top-5 accuracy evaluates whether the correct label appears among the five most confident predictions. In low-light action recognition, Top-5 accuracy is particularly informative because severe degradation may make several semantically similar actions difficult to separate from a single clip. Reporting both metrics therefore provides a more complete view of recognition performance.

\subsection{Comparison with State-of-the-Art on ELLAR}
Table~\ref{tab:ellar_sota} compares the proposed framework with previously reported methods on the ELLAR dataset. The proposed model achieves the best performance, reaching \textbf{55.13\%} Top-1 accuracy and \textbf{78.87\%} Top-5 accuracy on the test set. Compared with the reported DGAM result, this corresponds to an absolute gain of \textbf{16.71\%} in Top-1 and \textbf{4.43\%} in Top-5.These gains are substantial in the context of low-light HAR, where recognition is strongly affected by degraded appearance, weak motion boundaries, and unstable illumination. The results indicate that conditioning enhancement and classification on illumination characteristics is more effective than relying on a fixed or loosely coupled processing strategy. In particular, the improvement over DGAM suggests that the proposed design is able to preserve more discriminative spatio-temporal cues under severe low-light conditions.
\begin{table}[htbp]
\centering
\caption{Comparison with state-of-the-art methods on the ELLAR dataset.}
\label{tab:ellar_sota}
\scriptsize
\renewcommand{\arraystretch}{1.1}
\setlength{\tabcolsep}{4pt}
\begin{tabular}{lcccc}
\toprule
\textbf{Method} & \textbf{Pretrained} & \textbf{Input Setting} & \textbf{Top-1} & \textbf{Top-5} \\
\midrule
ResNet101 & K700 & RGB clip & 10.46 & 45.69 \\
ResNeXt101 & K400 & RGB clip & 9.63 & 39.37 \\
DarkLight & IG-65M & RGB clip & 28.58 & 64.31 \\
TimeSformer & K400 & RGB clip & 15.51 & 55.96 \\
Video-Swin-B & K400 & RGB clip & 35.03 & 68.87 \\
DGAM & K400 & RGB clip & 38.42 & 74.44 \\
\textbf{Illumi-Net}(Proposed) & \textbf{VideoMAE-B} & \textbf{RGB clip} & \textbf{55.13} & \textbf{78.87} \\
\bottomrule
\end{tabular}
\end{table}

\subsection{Benchmark Results on LUMINA-26}
We further evaluate the proposed framework on LUMINA-26, the real-world low-light action recognition dataset introduced in this work. The proposed method achieves \textbf{75.95\%} Top-1 accuracy and \textbf{93.58\%} Top-5 accuracy on the test set. These results indicate that the model can learn robust discriminative representations despite severe darkness, motion ambiguity, and substantial scene variation.
Since LUMINA-26 is introduced in this work, these results also serve as the first benchmark reference for future studies. The strong performance shows that the dataset is challenging yet learnable, and that the proposed Illumi-Net framework can effectively handle realistic night-time variability rather than only highly controlled or filtered low-light conditions.

\begin{table}[htbp]
\centering
\caption{Performance of the proposed method on the LUMINA-26 dataset.}
\label{tab:lumina_comparison}
\scriptsize
\renewcommand{\arraystretch}{1.1}
\setlength{\tabcolsep}{6pt}
\begin{tabular}{lcc}
\toprule
\textbf{Model} & \textbf{Top-1} & \textbf{Top-5} \\
\midrule
\textbf{Illumi-Net} & \textbf{75.95} & \textbf{93.58} \\
\bottomrule
\end{tabular}
\end{table}

\subsection{Ablation Study}
To quantify the contribution of each component, we conduct ablation experiments on the ELLAR dataset. Specifically, we analyze four configurations: backbone only, illumination enhancement only, illumination enhancement with adaptive gating, and the full model.

The ablation results in Table~\ref{tab:ablation_results} show a clear performance progression. The backbone-only model provides a strong baseline, while illumination enhancement alone is insufficient. Introducing adaptive gating further improves performance, and the full model achieves the best overall results. This confirms that the combination of illumination-aware enhancement, adaptive expert routing, and expert-conditioned classification is essential for robust low-light HAR.

\begin{table}[htbp]
\centering
\caption{Ablation results for the proposed framework on the ELLAR dataset.}
\label{tab:ablation_results}
\scriptsize
\renewcommand{\arraystretch}{1.1}
\setlength{\tabcolsep}{3pt}
\begin{tabular}{lcccccc}
\toprule
\textbf{Exp} & \textbf{Dataset}  & \textbf{Top-1} & \textbf{Top-5} & \textbf{Notes} \\
\midrule
A1\_backbone & ELLAR  & 0.528 & 0.768 & Backbone only \\
A2\_enhancement & ELLAR  & 0.455 & 0.737 & Illumination only \\
A3\_enh+gating & ELLAR  & 0.529 & 0.763 & Illumination + gating \\
A4\_full & ELLAR  & 0.543 & 0.778 & Full MoE model \\
\bottomrule
\end{tabular}
\end{table}

\subsection{Discussion of Findings}
Taken together, the experimental results support two main observations. First, the proposed framework is highly effective on an established low-light benchmark, as evidenced by the large margin over prior methods on ELLAR. Second, the same framework transfers successfully to LUMINA-26 and establishes a strong benchmark on a newly introduced real-world low-light dataset. Ablation studies show that performance gains arise from the coordinated interaction between illumination-aware enhancement, adaptive routing, and expert-conditioned classification, rather than from any single component. This highlights the importance of incorporating illumination awareness into both enhancement and classification stages for robust low-light human action recognition.

\section{Results and Discussion}
\label{sec:results_discussion}
The experimental results validate both central contributions of this work: the proposed LUMINA-26 dataset and the Illumi-Net: An Illumination-Adaptive Mixture-of-Experts Network. From the modeling perspective, the proposed method consistently improves recognition under severe low-light conditions. On the ELLAR benchmark, it surpasses the previously reported state-of-the-art DGAM result by a clear margin in both Top-1 and Top-5 accuracy. This improvement indicates that combining illumination-aware enhancement with adaptive expert selection is more effective than relying on a fixed low-light correction strategy. Rather than processing all dark videos in the same way, the proposed framework adapts its response to the visual characteristics of each input, making it more robust to variations in darkness level, noise, blur, and scene complexity.
From the dataset perspective, LUMINA-26 provides a practically meaningful benchmark for low-light human action recognition. Compared with earlier low-light HAR datasets, it includes a larger action vocabulary while maintaining strong class balance and substantial real-world capture diversity. This is important because benchmarks with broader action categories, more subjects, and richer scene variation provide a more realistic measure of generalization beyond controlled settings. The strong performance of the proposed model on LUMINA-26 further shows that the dataset is challenging yet learnable, making it suitable for future benchmarking and method development.
The ablation study on ELLAR provides additional insight into why the proposed framework is effective. The backbone-only configuration already offers a strong baseline, suggesting that transformer-based video models retain useful spatio-temporal information even under degraded visibility. However, the enhancement-only configuration does not provide the best performance, which indicates that simple illumination correction alone is insufficient. In low-light HAR, making a video brighter does not necessarily make the action easier to classify, since naive enhancement can also amplify noise or distort subtle motion cues. When adaptive gating is introduced, performance improves further, showing that the combination of illumination-aware enhancement and expert-conditioned routing enables the model to preserve discriminative spatio-temporal features under varying low-light conditions. The full model achieves the highest Top-1 and Top-5 accuracy, confirming that each component contributes meaningfully to the overall performance and that coordinated adaptation across enhancement and classification is critical for robust low-light HAR.
Overall, these results demonstrate that both the dataset and the proposed framework complement each other: LUMINA-26 provides challenging yet realistic data, and the Illumi-Net framework effectively exploits this data to achieve state-of-the-art performance in low-light human action recognition.

\section{Conclusion}
\label{sec:conclusion}

In this work, we introduced LUMINA-26, a real-world low-light human action recognition dataset designed to support more realistic benchmarking under dark and night-time conditions. With 26 action classes and substantial diversity in subjects, recording locations, and visual conditions, the dataset offers a more practical and representative benchmark than earlier low-light HAR datasets with limited action vocabulary or more constrained capture settings. Along with the dataset, we proposed Illumi-Net: An Illumination-Adaptive Mixture-of-Experts Network that integrates enhancement, adaptive routing, and expert-conditioned classification within a unified architecture.
Extensive experiments demonstrated the effectiveness of the proposed framework. The model achieved strong performance on LUMINA-26 and also outperformed the previously reported state-of-the-art result on the ELLAR benchmark. In addition, the ablation study showed that the observed gains are not due to simple brightness correction alone, but arise from the coordinated interaction between illumination-aware enhancement, adaptive expert routing, and expert-conditioned recognition.
The findings of this study highlight two broader points. First, low-light human action recognition remains a challenging problem because real-world night-time videos are affected by complex variations in darkness, noise, blur, and scene structure. Second, progress in this area depends not only on stronger recognition models, but also on realistic and well-designed benchmarks that better reflect practical deployment conditions. We hope that LUMINA-26 will serve as a useful resource for future research and encourage further work on robust, adaptive, and generalizable low-light HAR methods.
In future work, we plan to extend the benchmark with more challenging evaluation protocols and broader cross-domain testing, and to investigate more efficient low-light adaptive architectures for real-time applications such as surveillance, assistive systems, and night-time scene understanding.

\section*{Acknowledgement}

The authors gratefully acknowledge the 22 students from the Machine Learning Lab, Delhi Technological University, and the Department of Computer Science, Delhi College of Arts and Commerce, University of Delhi, for their support in recording the LUMINA-26 dataset. Their active participation in action performance, recording coordination, and data collection under real-world low-light conditions was essential to this study.
\bibliographystyle{model1-num-names.bst}
\bibliography{lumina}

\end{document}